\documentclass[letterpaper, 10 pt, conference]{ieeeconf} 

\IEEEoverridecommandlockouts   
\usepackage{graphicx} 
\usepackage{amsmath} 
\usepackage{amssymb} 
\usepackage{booktabs} 
\usepackage{siunitx} 
\usepackage{color} 
\usepackage{cite}
\usepackage{subcaption}
\usepackage{graphicx}
\usepackage{makecell}
\usepackage{algorithm, float}
\usepackage{algpseudocode}

\usepackage[hidelinks]{hyperref}
\title{\LARGE \bf
A Deployable, Bio-inspired Compliant Leg Design for Enhanced Leaping in Quadruped Robots
}

\author{Yiyang Chen$^{1\dagger}$, Yuxin Liu$^{1\dagger}$, Jinzheng Zhou$^{1\dagger}$, Fanxin Wang$^{1*}$, Qinglei Bu$^{1}$, Jie Sun$^{1}$ and Yikun Cheng$^{2}$
\thanks{$^{\dagger}$These authors contributed equally to this work.}
\thanks{$^{1}$Yiyang Chen, Yuxin Liu, Jinzheng Zhou, Fanxin Wang, Qinglei Bu, and Jie Sun are with the Department of Mechatronics and Robotics, Xi'an Jiaotong-Liverpool University, Suzhou 215123, China.}
\thanks{$^{2}$Yikun Cheng is with the Department of Mechanical Science and Engineering, University of Illinois at Urbana-Champaign, Champaign, Illinois, United States.}
\thanks{*Correspondence: Fanxin Wang
{\tt\small Fanxin.Wang@xjtlu.edu.cn}}%
}

\begin{document}
\maketitle
\thispagestyle{empty} 
\pagestyle{empty} 

\begin{abstract}

Quadruped robots are becoming increasingly essential for various applications, including industrial inspection and catastrophe search and rescue. These scenarios require robots to possess enhanced agility and obstacle-navigation skills. Nonetheless, the performance of current platforms is often constrained by insufficient peak motor power, limiting their ability to perform explosive jumps. To address this challenge, this paper proposes a bio-inspired method that emulates the energy-storage mechanism found in froghopper legs. We designed a Deployable Compliant Leg (DCL) utilizing a specialized 3D-printed elastic material, Polyether block amide (PEBA), featuring a lightweight internal lattice structure. This structure functions analogously to biological tendons, storing elastic energy during the robot's squatting phase and rapidly releasing it to augment motor output during the leap. The proposed mechanical design significantly enhances the robot's vertical jumping capability. Through finite element analysis (FEA) and experimental validation, we demonstrate a relative performance improvement of \textbf{\SI{17.1} {\percent}} in vertical jumping height.

\end{abstract}

\section{INTRODUCTION}

Quadruped robots have demonstrated impressive performance in diverse applications, such as industrial inspection and emergency response~\cite{fan2024review, lee2024safety, li2023fabrication}. Their legged design offers significant advantages over wheeled systems in navigating unstructured and complex terrains. However, enhancing their explosive motion capabilities, particularly jumping, remains a key challenge. Jumping requires rapid velocity generation within a short horizon, which is fundamentally limited by physical constraints such as actuator torque capacity, joint kinematic limits, and ground friction cone constraints~\cite{wang2023research}.

\begin{figure}[ht!]
  \centering
  \includegraphics[width=0.9\columnwidth]{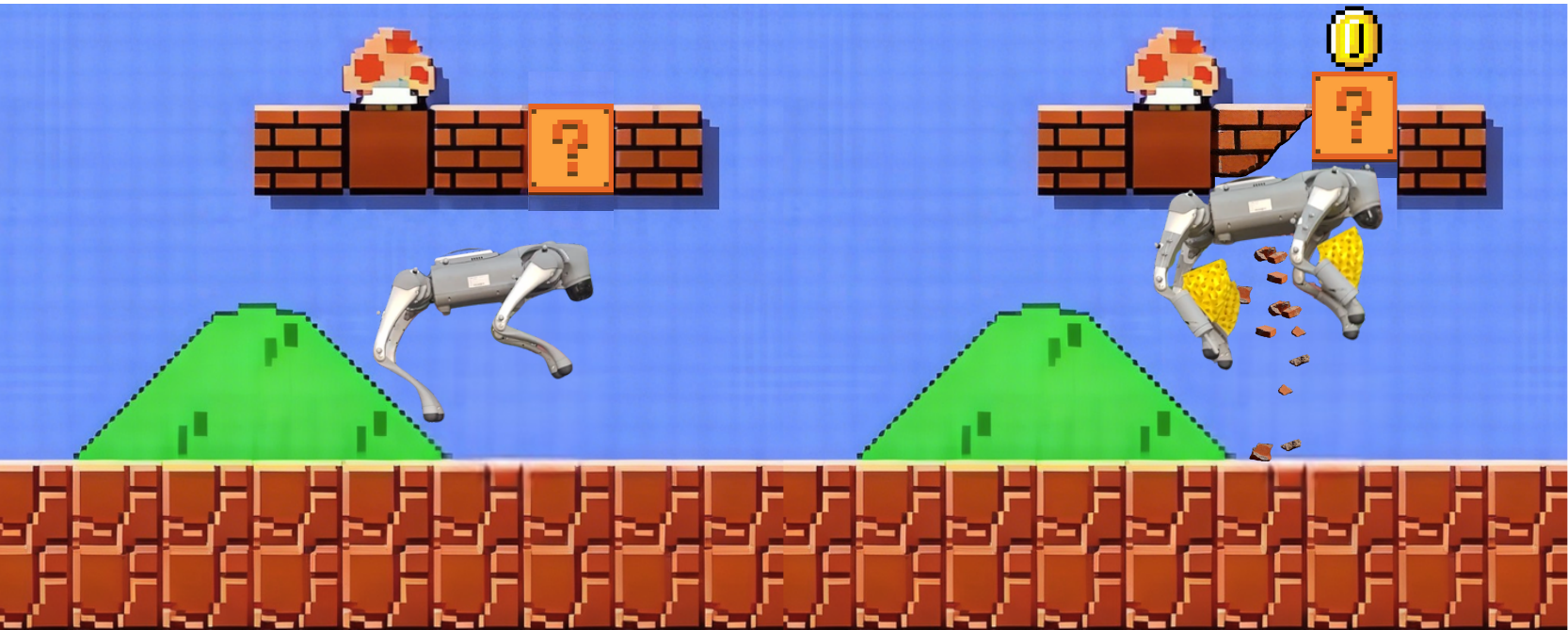} 
  
  \caption{\textbf{Illustration of the Jumping Actuation Challenge.} \\
  \textbf{(Left)} The baseline robot, limited by motor torque saturation, fails to reach the critical height. \\
  \textbf{(Right)} The proposed system, augmented with DCLs, releases extra stored elastic energy and completes the task (symbolized by the coin).}
  
  \label{fig:mario_concept}
\end{figure}

Although current quadruped robots driven by electric motors can deliver strong jumping performance, they heavily depend on rigid-body dynamics and high-torque joint actuators, particularly Direct-Drive and Quasi-Direct-Drive motors~\cite{ye2021modeling,  luo2022prismatic, kau2019stanford}, whose torque-density performance is at or approaching the practical limits~\cite{ye2021modeling}. Dynamic motions, particularly leaping, are crucial for overcoming significant obstacles and improving overall agility. Conventional industrial systems are able to apply high-rated power actuators. By contrast, legged robots frequently operate in dynamic and complex environments that require rapid motion response~\cite{meng2022explosive}. The primary limitation stems from the challenge of generating sufficient explosive actuation, often requiring motors with high peak torque, which adds weight and cost. This "actuation bottleneck" is conceptually illustrated in Fig. \ref{fig:mario_concept}, where the baseline system fails to overcome the potential energy barrier. Furthermore, precise control and coordination are necessary to manage the high-impact forces involved.

To address this, researchers have explored several avenues. One approach involves applying parallel elastic elements to enhance power density. Recent studies have demonstrated that articulated soft structures~\cite{apostolides2025explosive} and optimized parallel elastic actuators~\cite{liu2024novel} can significantly improve explosive jumping performance by buffering impact and releasing stored energy. A more bio-inspired approach utilizes passive compliant elements to store and release energy, mimicking biological systems. For instance, recent works in mechatronics have shown the value of compliance for creating adaptable locomotion through morphing structures, such as compliant actuators that imitate biological muscle~\cite{yang2025compliant}.

\begin{figure*}[thpb]
\centering
\includegraphics[width=0.72\textwidth]{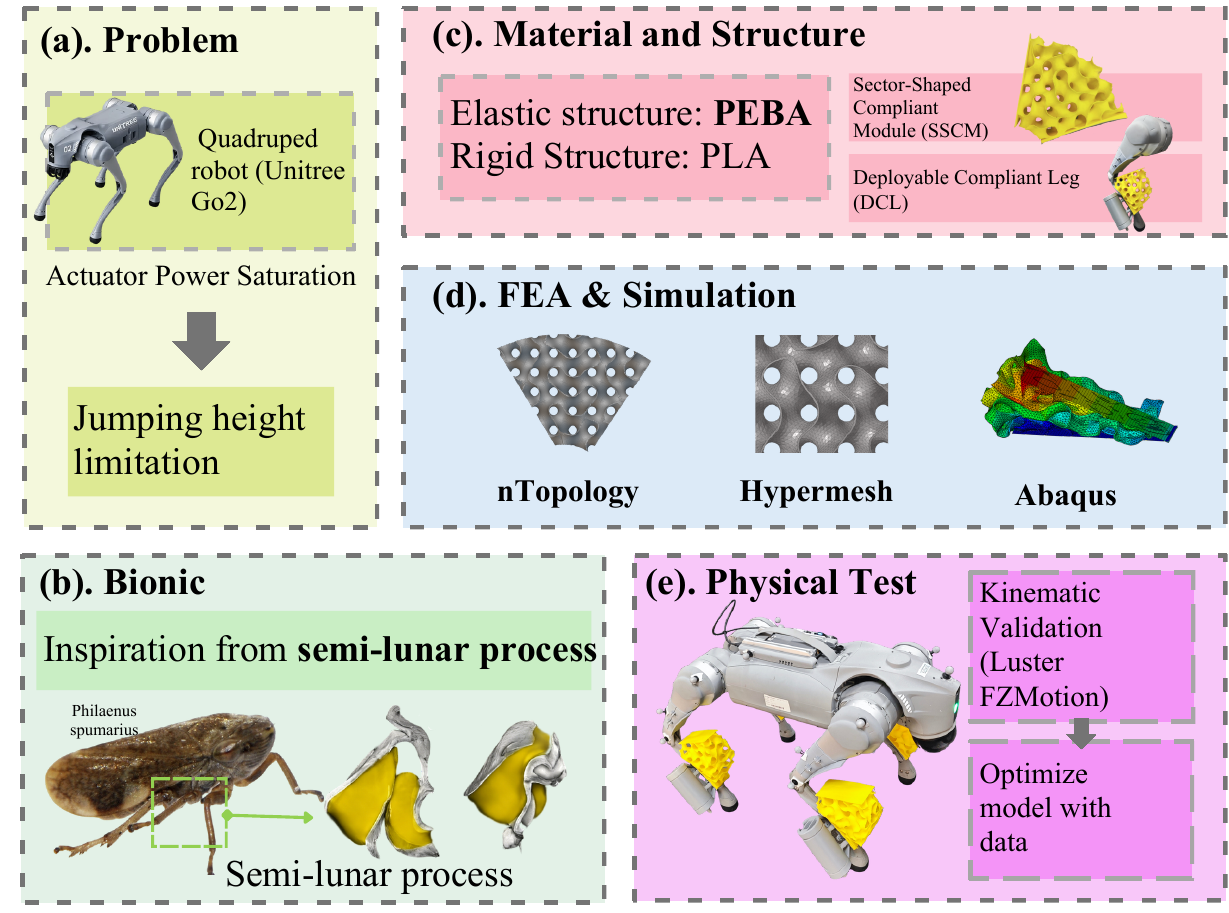}
\caption{(a-e) Design workflow of the bio-inspired quadruped leaping enhancement; (b) The biological prototype \cite{sutton2018insect}.}
\label{fig:cad_overview}
\end{figure*}


However, existing hardware solutions, involving passive compliance, present a fundamental mechanical trade-off. 
While Parallel Elastic Actuators (PEAs) and soft structures effectively enhance peak power density~\cite{liu2024novel, apostolides2025explosive}, they are typically permanently integrated into the robot's kinematic chain. 
This constant engagement creates a critical issue during nominal locomotion: the passive elasticity acts as \textbf{parasitic stiffness} that the motors must continuously overcome.
This parasitic effect not only increases the Cost of Transport (CoT) but also limits the bandwidth of joint control during non-explosive tasks. 
Although Variable Stiffness Actuators (VSAs) can theoretically address this by adjusting compliance, they often require complex transmission mechanisms and additional heavy motors~\cite{10081107}, rendering them unsuitable for compact, lightweight quadruped legs. 
Consequently, there remains a lack of practical structural designs that can offer high-density energy storage on demand without compromising the efficiency of the robot's primary locomotion gaits.

In this work, we refer to the entire leg assembly as the Deployable Compliant Leg (DCL), which is integrated with a specific energy-storage component termed the Sector-Shaped Compliant Module (SSCM). The entire design workflow of DCL is shown in Fig. \ref{fig:cad_overview}. The primary contributions of this work are:
\begin{itemize}
\item \textbf{Innovative Structural Design:} 
A bio-inspired DCL that integrates an optimized Gyroid lattice structure for efficient energy storage. By utilizing the high-performance material PEBA, the SSCM is capable of storing substantial elastic energy.
\item \textbf{Deployable Flipping Mechanism:} 
A novel flipping mechanism enabling efficient and rapid reconfiguration between an energy‑efficient walking configuration (stowed state) and a high‑amplitude leaping configuration (deployed state), thereby enhancing the dynamic agility of legged robotic systems.
\item \textbf{Integrated Modeling and Testing:}
A comprehensive integrated design-to-fabrication framework that combines structural optimization, FEA parameterization, and real-world validation, achieving a \textbf{17.1\%} increase in vertical leaping height.

\end{itemize}

The remainder of this paper is organized as follows: Section II details the design of the DCL. Section III describes the FEA simulation, characterization, and design optimization of the structure. Section IV presents the experimental validation and results. Finally, Section V concludes the paper and discusses future work. A video demonstrating our work could be found in: \textcolor{blue}{\href{url}{https://b23.tv/lj5qQBw}}.

\section{DESIGN OF THE DEPLOYABLE COMPLIANT LEG}

This section details the mechanical architecture and design methodology of the proposed deployable compliant leg. As conceptualized in Fig.\ref{fig:mario_concept}, the system integrates a bio-inspired energy-storage module with a Deployable Flipping Mechanism to realize two primary functional objectives: high-density energy release for explosive leaping and on-demand reconfiguration to eliminate parasitic stiffness during nominal locomotion.

\subsection{Bio-inspired Concept and Overall Architecture}
\begin{figure*}[t] 
    \centering
    \includegraphics[width=0.6\linewidth]{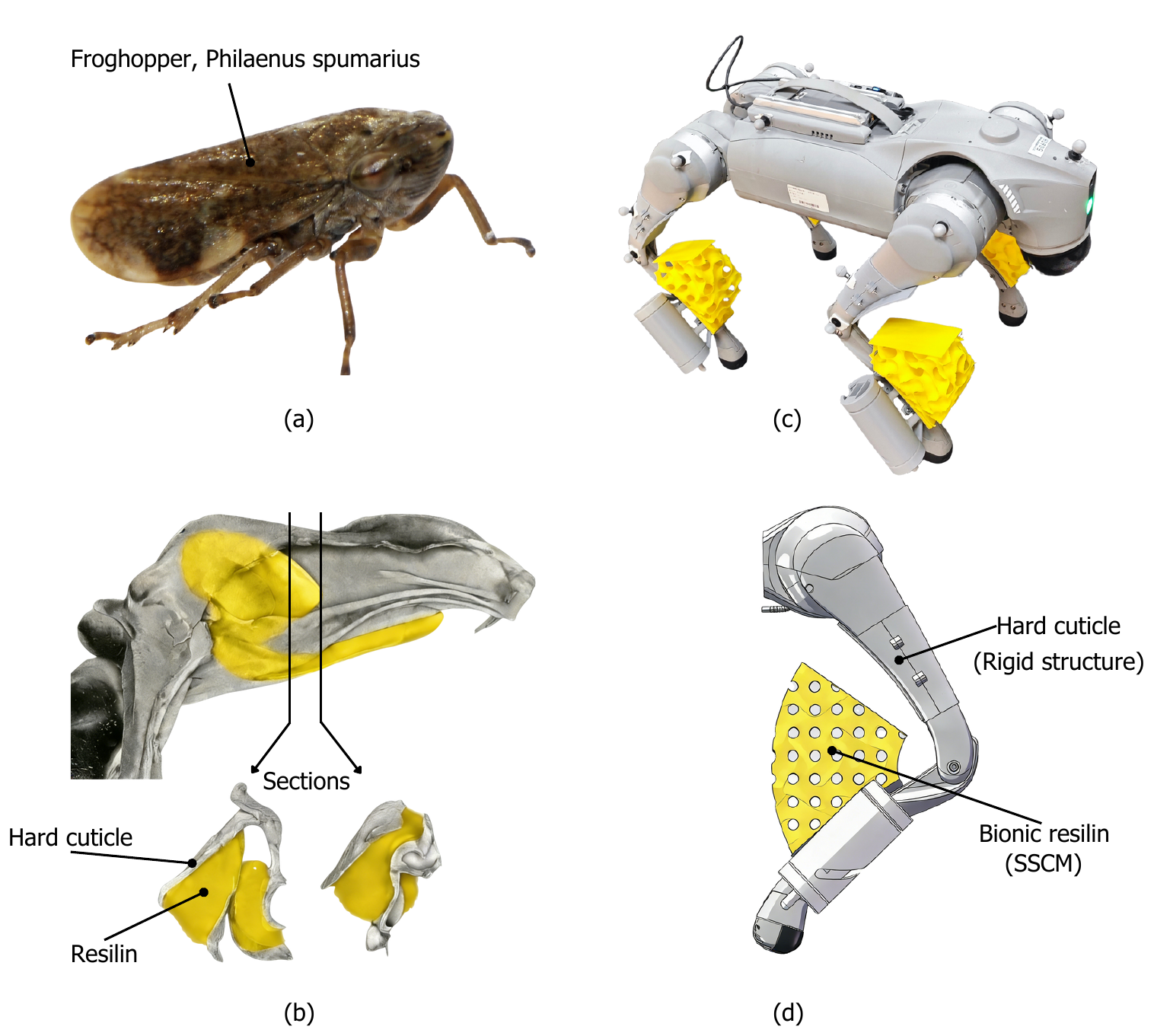}
    
    \caption{Bio-inspired design concept of the robotic leg. 
    (a) The Philaenus spumarius, serving as the biological prototype \cite{sutton2018insect}. 
    (b) Biological mechanism highlighting the interaction between the rigid hard cuticle and the energy-storing resilin \cite{sutton2018insect}. 
    (c) The proposed quadruped robot integrated with the SSCMs. 
    (d) Mechanical design overview.}
    
    \label{fig:bio_design}
\end{figure*}

 The design draws inspiration from the biomechanics of the froghopper (\textit{Philaenus spumarius} Fig. \ref{fig:bio_design}a), renowned for its explosive jumping capability. 
Unlike mammals that rely primarily on muscle power, these insects utilize a catapult mechanism powered by the \textbf{semi-lunar process} (or pleural arch) \cite{sutton2018insect}. 
As shown in Fig. \ref{fig:bio_design}b, this composite structure—consisting of the elastic protein resilin (yellow) and rigid cuticle (grey)—forms a distinct \textbf{curved, sector-like geometry} around the trochanteral joint.

Crucially, this sector morphology is not merely structural but functional. 
During the squatting phase (flexion), the arch bends, storing elastic potential energy in the resilin through material compression and structural buckling. 
This geometry provides a non-linear stiffness profile: it is compliant initially to allow flexion but stiffens progressively to store high-density energy, perfectly matching the kinematic constraints of the leg joint \cite{burrows2012locusts}.

We implement this approach by creating an SSCM that interfaces with the robot's limb, as depicted in the integrated system view (Fig. \ref{fig:bio_design}c). To form the SSCM, Arkema Pebax\textsuperscript{\textregistered} 3533 SP 01 is used to serve as the Bionic Resilin (SSCM), and PLA serves as the Hard Cuticle (Rigid Structure), with the specific structural layout detailed in Fig. \ref{fig:bio_design}d. Energy is 
stored during the squatting phase of the quadruped robot and during jumping. The arch structure forms during the jumping process and plays a similar role to the semi-lunar process in the froghopper.

\subsection{Material Selection}
The structure is fabricated from Arkema Pebax\textsuperscript{\textregistered} 3533 SP 01, a thermoplastic elastomer selected for its high elasticity, durability, and low weight. These characteristics make it ideal for rapid prototyping elastic supporting structures that emulate the mechanical properties of resilin in froghoppers. Thermoplastic elastomers (TPEs) are chosen to use as the elastic structure, and there are four primary categories of commercial TPE materials: polystyrene elastomer, polyurethane elastomer, polyester elastomer, and polyamide elastomer \cite{honeker1996impact}. PEBA is a type of TPE that provides a distinctive blend of lightweight, flexibility, minimal hysteresis, and outstanding resistance to flex fatigue \cite{wang2018lightweight}. It is made of polyamide (hard segment) and rubbery polyether (soft segment). The hard segments act as physical crosslinks, enhancing hardness and stiffness, while the soft segments provide flexibility and elasticity. Adjusting the ratio of these segments allows for tailored hardness of PEBA \cite{xu2022super}. In comparison to traditional thermoplastic elastomers like thermoplastic polyolefin elastomer (TPO) and thermoplastic polyurethane (TPU), PEBA demonstrates a reduced material density and energy loss factor. 
The low energy loss factor and minimal hysteresis make PEBA capable of serving as a Bionic Resilin, releasing energy during the jumping phase, like a spring. The low density of PEBA minimizes the mass penalty on the robot, while its exceptional fatigue resistance ensures a prolonged service life.

\subsection{SSCM and Lattice Structure}
To replicate the bio-mechanical functionality of the semi-lunar process, the \textbf{SSCM} design is proposed in this section. 
Unlike traditional cylindrical springs, this sector geometry is optimized to fit the angular space formed between the robot's shank and thigh during the deep squat (approx. $30^{\circ}$ to $90^{\circ}$). 
This configuration serves two critical functions:
1) \textbf{Geometric Compatibility}: It maximizes the volume available for the energy-absorbing lattice within the compact joint constraints.
2) \textbf{Progressive Stiffness}: The sector shape undergoes non-uniform compression—the inner radius compresses more densely than the outer radius—creating a non-linear stiffness profile that mimics the biological prototype, preventing premature bottoming-out while ensuring explosive energy release.


To precisely control the mechanical properties and maximize specific energy absorption, we utilized nTopology (nTop) to generate and optimize the internal lattice structure. Among the Triply Periodic Minimal Surface (TPMS) family, we conducted a comparative analysis of Diamond, Schwarz Primitive, Lidinoid, and Gyroid architectures.

\begin{figure}[h]  
    \centering
    \begin{subfigure}[b]{0.21\textwidth}  
        \centering
        \includegraphics[width=\textwidth]{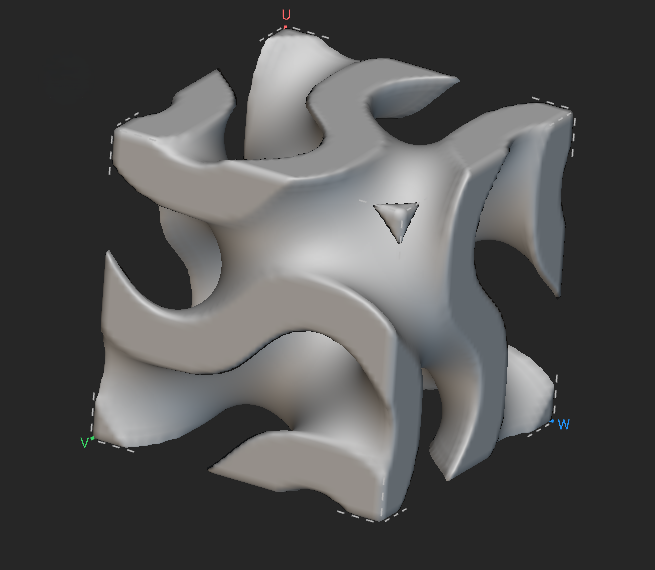}
        \subcaption*{(A)}
        \label{fig:left_part}
    \end{subfigure}
    \hfill 
    \begin{subfigure}[b]{0.203\textwidth}
        \centering
        \includegraphics[width=\textwidth]{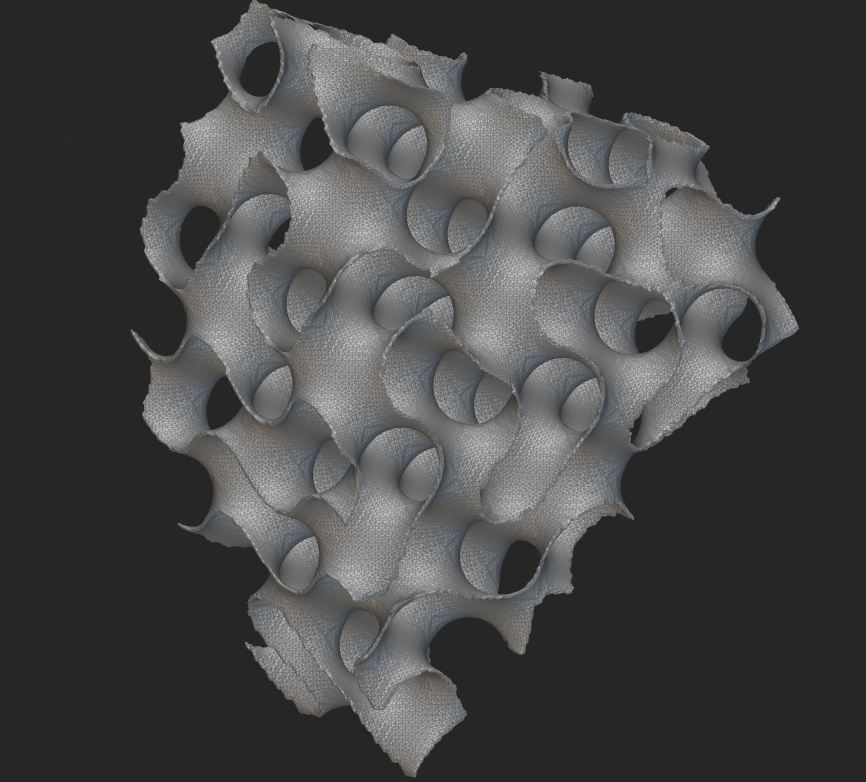}
        \subcaption*{(B)}
        \label{fig:right_part}
    \end{subfigure}
    
    \caption{The Gyroid lattice structure designed in nTop, shown (A) as a unit cell and (B) implemented as the internal filler of the compliant structure.}
    \label{fig:lattice}
\end{figure}

While Diamond structures offer high stiffness, they are prone to stress concentrations at nodal junctions. In contrast, the Gyroid structure—defined by the equation $\sin x \cos y + \sin y \cos z + \sin z \cos x = t$, was selected for its constant mean curvature and self-supporting characteristics beneficial for additive manufacturing (as shown in Fig. \ref{fig:lattice}). Recent studies ~\cite{lyu2024study, ahmad2025recent} indicate that the Gyroid lattice provides superior isotropy and structural stability across varying load vectors. These characteristics reduce the complexity of testing the performance of the structure under diverse angles.

Gyroid lattice also have its own manufacturing advantages, Maconachie et al.\cite{maconachie2020compressive} demonstrate that the gyroid matrix phase geometry offers a highly manufacturable structure, particularly with an increased number of cells. The specimens examined in this study showed only minor flaws, such as stair-stepping. Their research also shows that Lattice structures featuring gyroid unit cells offer numerous potential benefits compared to alternative cell topologies, including their structural resemblance to bone, which is advantageous for biomedical uses, and enhanced manufacturability due to the continuously varying inclination angles of the struts \cite{maconachie2020compressive}. Such performance reduces the difficulty in manufacturing, particularly for the 3D printing process used in our prototype. Support structures are not required within the lattice voids since the geometry is self-supporting, which improves printing quality and reduces post-processing workload.

\subsection{Deployable Flipping Mechanism}
\label{subsec:mechanism}

\begin{figure}[tb]
    \centering
    \begin{subfigure}[b]{0.48\columnwidth}
        \centering
        \includegraphics[width=\textwidth]{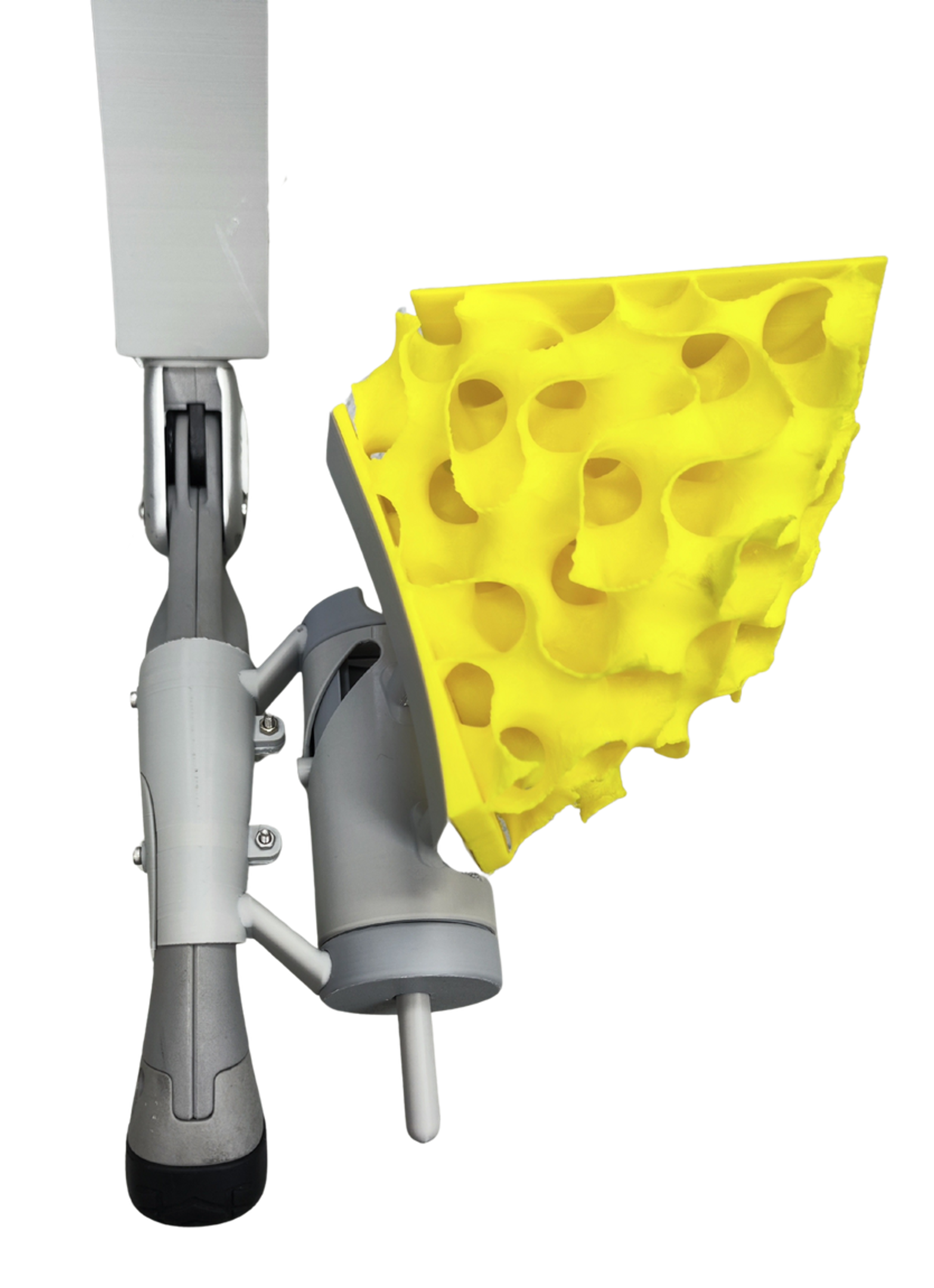} 
        \caption{Stowed State}
        \label{fig:stowed}
    \end{subfigure}
    \hfill 
    \begin{subfigure}[b]{0.39\columnwidth}
        \centering
        \includegraphics[width=\textwidth]{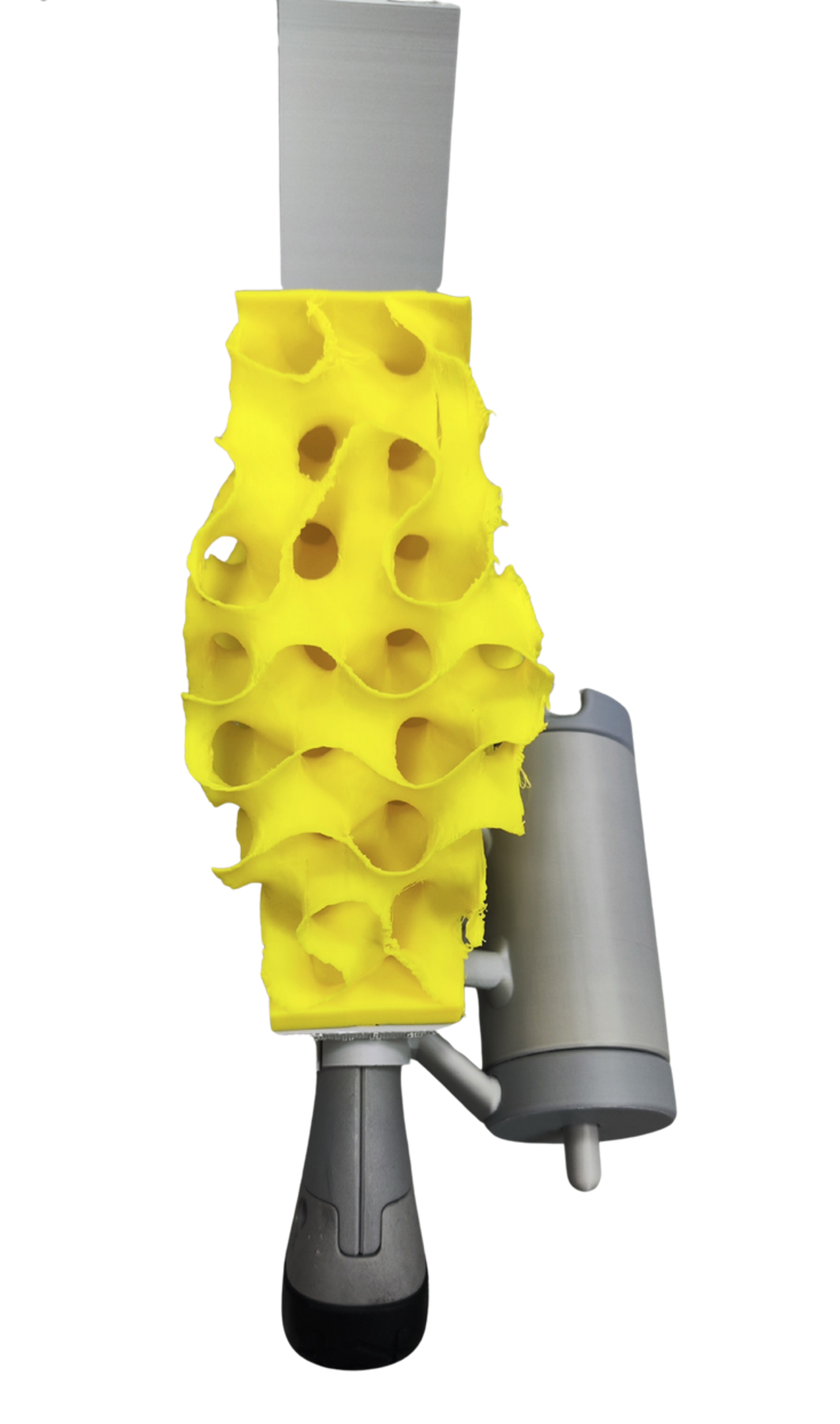} 
        \caption{Deployed State}
        \label{fig:deployed}
    \end{subfigure}
    \caption{\textbf{Operational states of the deployable mechanism.} (a) In the Stowed State, the mechanism retracts to eliminate parasitic stiffness. (b) In the Deployed State, the module rotates 90 degrees to align with the femur for ballistic energy storage.}
    \label{fig:states_comparison}
\end{figure}

To address the critical issue of parasitic stiffness during nominal locomotion, we engineered a \textbf{bistable Deployable Flipping Mechanism}. As illustrated in Fig. \ref{fig:states_comparison}, this compact assembly enables the Sector-Shaped Compliant Module (SSCM) to toggle between two distinct operational modes:
\begin{itemize}
    \item \textbf{Stowed State (Fig. \ref{fig:states_comparison}a):} The module is retracted against the lateral side of  the robot's calf, providing necessary clearance to prevent interference with the ground or the thigh during standard walking gaits.
    \item \textbf{Deployed State (Fig. \ref{fig:states_comparison}b):} The module rotates $90^{\circ}$ to align with the thigh, acting as a hard stop to engage the PEBA lattice for energy storage during the deep squat phase.
\end{itemize}

\begin{figure}[tb]
    \centering
    \includegraphics[width=0.7\columnwidth]{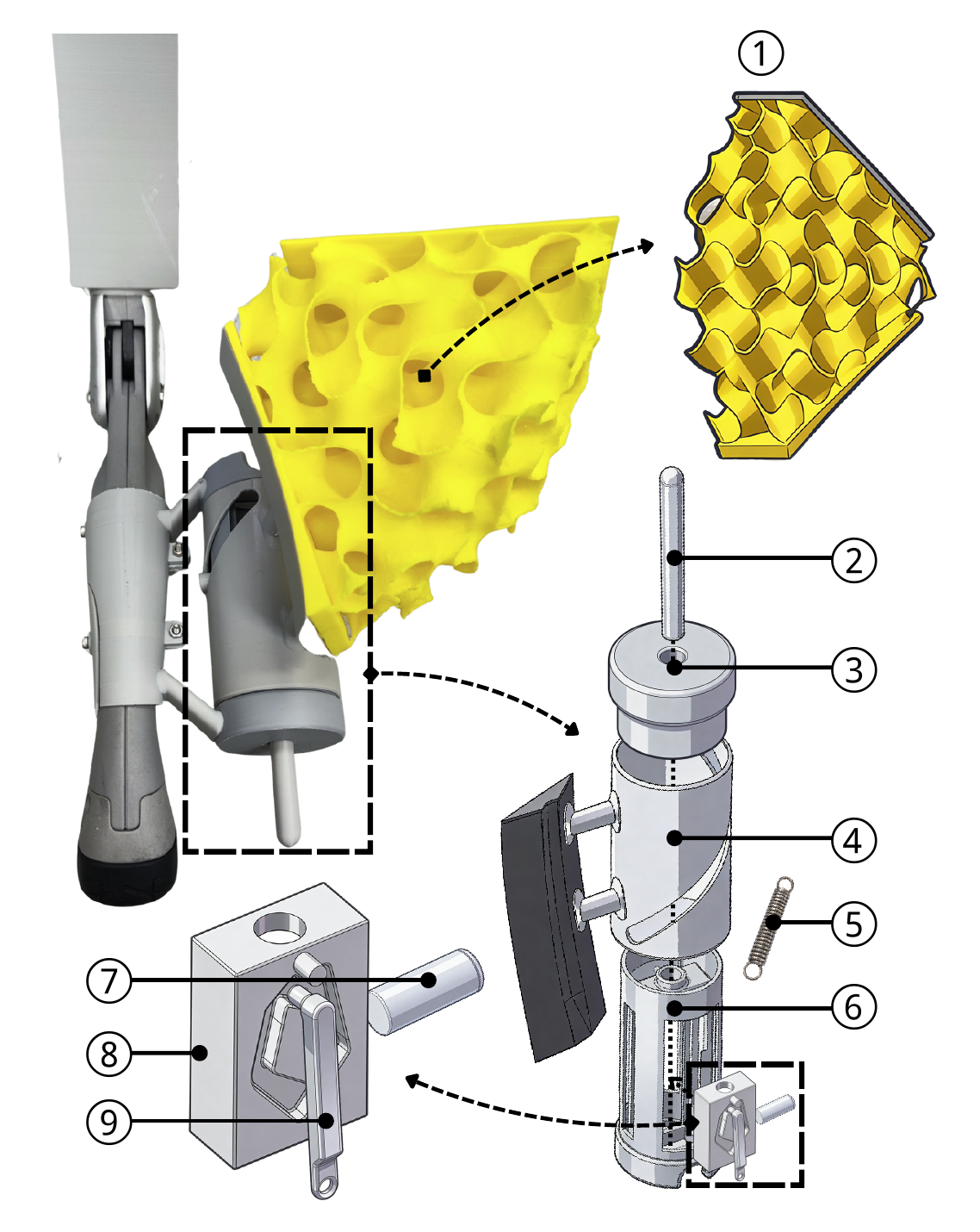} 
    \caption{\textbf{Exploded view of the kinematic chain.} Key components: (1) SSCM Lattice, (2) Push Rod, (3) Top Cap, (4) Rotating Sleeve, (5) Return Spring, (6) Inner Fixed Post, (7) Guide Pin, (8) Sliding Block, and (9) Locking Pin. The mechanism converts linear input into a precise 90-degree rotation via a helical cam constraint.}
    \label{fig:flip_mechanism}
\end{figure}

To achieve this complex spatial reconfiguration within a compact volume, the mechanism operates as a spatial cam system driven by linear actuation. The internal kinematic chain is detailed in the exploded view in Fig. \ref{fig:flip_mechanism}.

The motion transmission relies on the coupling of vertical translation and helical rotation:
\begin{enumerate}
    \item \textbf{Vertical Drive:} The external actuation acts directly on the \textbf{Push Rod (2)}, which in turn impels the \textbf{Sliding Block (8)}. Constrained by the \textbf{Inner Fixed Post (6)}, the Sliding Block is forced to translate vertically.
    
    \item \textbf{Rotational Conversion:} As the Sliding Block translates, the \textbf{Guide Pin (7)} mounted on its lateral side moves synchronously. This linear motion forces the \textbf{Outer Rotating Sleeve (4)} to rotate via the helical cam constraint, converting the vertical input into the required angular output.
\end{enumerate}


A \textbf{Return Spring (5)} maintains system tension, ensuring the internal \textbf{Locking Pin (9)} snaps into designated detents. This design creates a bistable feature, mechanically locking the system in either the stowed or deployed state without requiring continuous actuation torque.


\section{FEA AND STIFFNESS CHARACTERIZATION}

\subsection{Finite Element Formulation}
To validate the structural integrity of the lattice under large deformations and derive the stiffness model for control, a high-fidelity finite element model was established in \textbf{Abaqus/Explicit}.

\subsubsection{Constitutive Modeling and Discretization}
The Sector-Shaped Compliant Module (SSCM) is composed of Arkema Pebax 3533 SP 01. To accurately capture its hyperelastic response, a \textbf{Marlow potential} was constructed based on the manufacturer's uniaxial test data~\cite{arkema_tds}. This constitutive model was selected to strictly reproduce the experimental stress-strain response, avoiding curve-fitting oscillations often associated with polynomial-based potentials (e.g., Yeoh) at high strains. A Poisson's ratio of 0.48 was applied to model the material's near-incompressibility while preventing volumetric locking.

The geometry was discretized using modified quadratic tetrahedral elements (C3D10M). This element formulation provides superior robustness against mesh distortion and contact instabilities under large deformation compared to linear approximations.

\subsubsection{Boundary Conditions and Simulation Setup}
A kinematic coupling was established at the knee's center of rotation to replicate the physical joint mechanics of the Unitree Go2. 
While the nominal operating range for a standard jump involves approximately $29^{\circ}$ of compression, the simulation was extended to $45^{\circ}$. This extended range ensures the characterization covers the \textit{Safety Margin} and identifies the onset of structural densification, verifying mechanical reliability under limit conditions.

\subsection{Stiffness Modeling for Control Integration}

\subsubsection{Torque-Deformation Characterization}
The FEA results (Fig.~\ref{fig:stiffness_curve}) reveal a distinct non-linear stiffening behavior. To facilitate control integration, the mechanical response is segmented into two domains: the \textit{Operating Region} ($0^{\circ} < \theta \le 29^{\circ}$), where dynamic leaping occurs, and the \textit{Safety Margin} ($29^{\circ} < \theta < 39^{\circ}$), acting as a buffer before the \textit{Densification Region} ($> 39^{\circ}$).

\subsubsection{Analytical Model Identification}
For real-time feedforward control, we developed a high-fidelity analytical model. Unlike global fitting approaches that may compromise local accuracy, we adopted a \textbf{domain-specific fitting strategy} confined strictly to the \textit{Operating Region}. The torque response was approximated using a \textbf{third-order} polynomial:
\begin{equation}
\tau_{\text{exo}}(\theta) = \alpha_3 \theta^3 + \alpha_2 \theta^2 + \alpha_1 \theta + \alpha_0
\label{eq:poly_fit}
\end{equation}
where $\alpha_i$ are the identified stiffness coefficients.

\begin{figure}[htbp]
    \centering
    \includegraphics[width=0.95\linewidth]{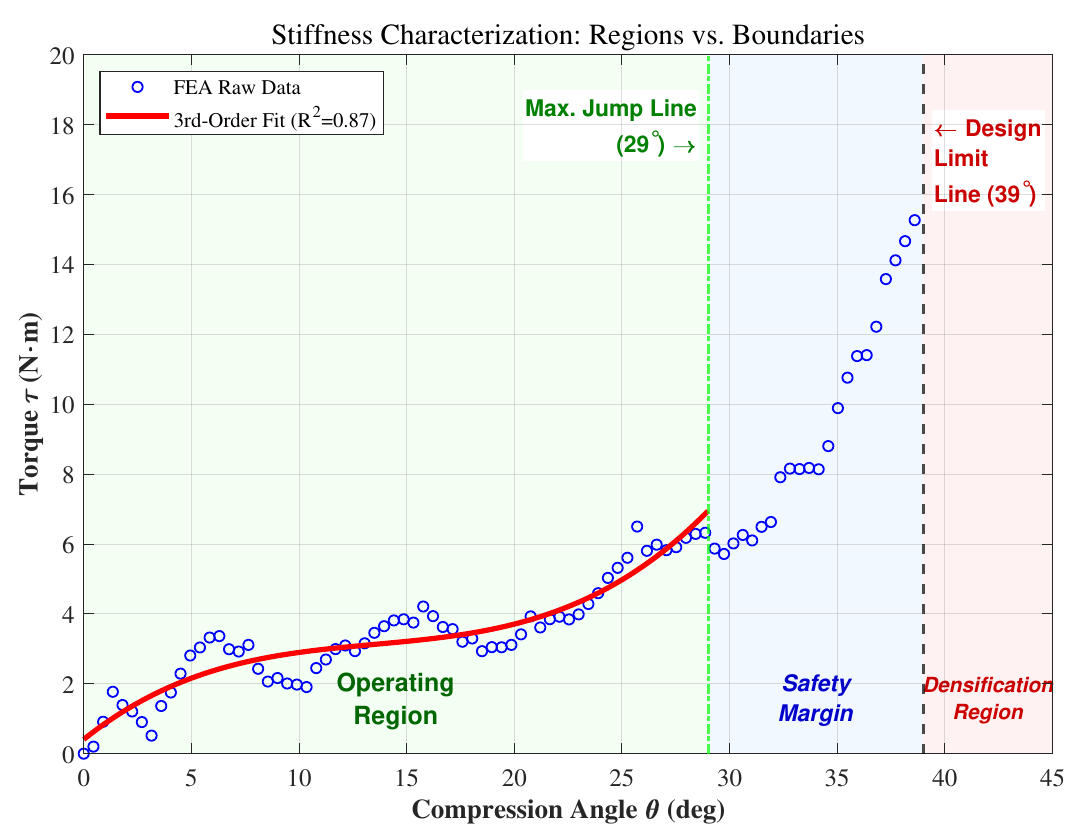} 
    \caption{Rigorous stiffness characterization and operating bounds. FEA raw data (blue circles) validates the structural integrity up to $45^{\circ}$. The \textbf{Analytical Model} (solid red line) is a $3^{rd}$-order polynomial fitted to the \textbf{Operating Region} ($0^{\circ}$--$29^{\circ}$) with maximize torque $6.8 N\cdot m$. The \textbf{Safety Margin} (blue zone) provides a $\approx 10^{\circ}$ buffer before the \textbf{Design Limit Line} (red, $39^{\circ}$), beyond which the structure enters the \textbf{Densification Region} (red zone).}
    \label{fig:stiffness_curve}
\end{figure}

As shown in Fig.~\ref{fig:stiffness_curve}, the model demonstrates excellent fidelity ($R^2 \approx 0.87$) within the operating bounds. Beyond the operating range, the FEA results confirm that the structure maintains a monotonically increasing stiffness profile within the Safety Margin, preventing abrupt bottoming-out.

\section{EXPERIMENT VALIDATION}

\begin{figure}[b!]
    \centering
    \begin{subfigure}[b]{0.48\linewidth}
        \centering
        \includegraphics[height=3.2cm]{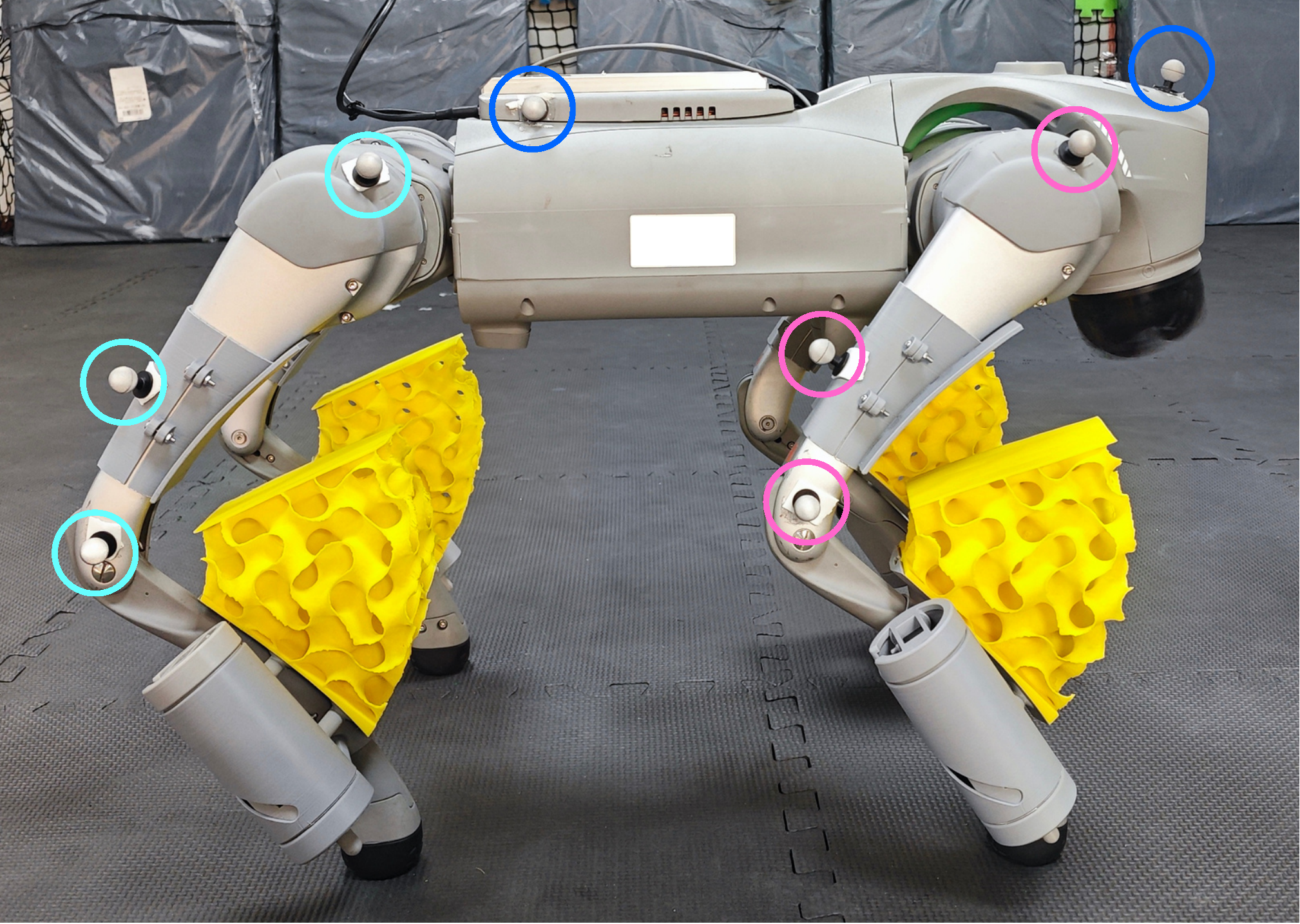}
        \subcaption*{(A)} 
        \label{fig:robot_setup}
    \end{subfigure}
    \hfill 
    \begin{subfigure}[b]{0.48\linewidth}
        \centering
        \includegraphics[height=3.2cm, trim={1cm 0cm 1cm 0cm}, clip]{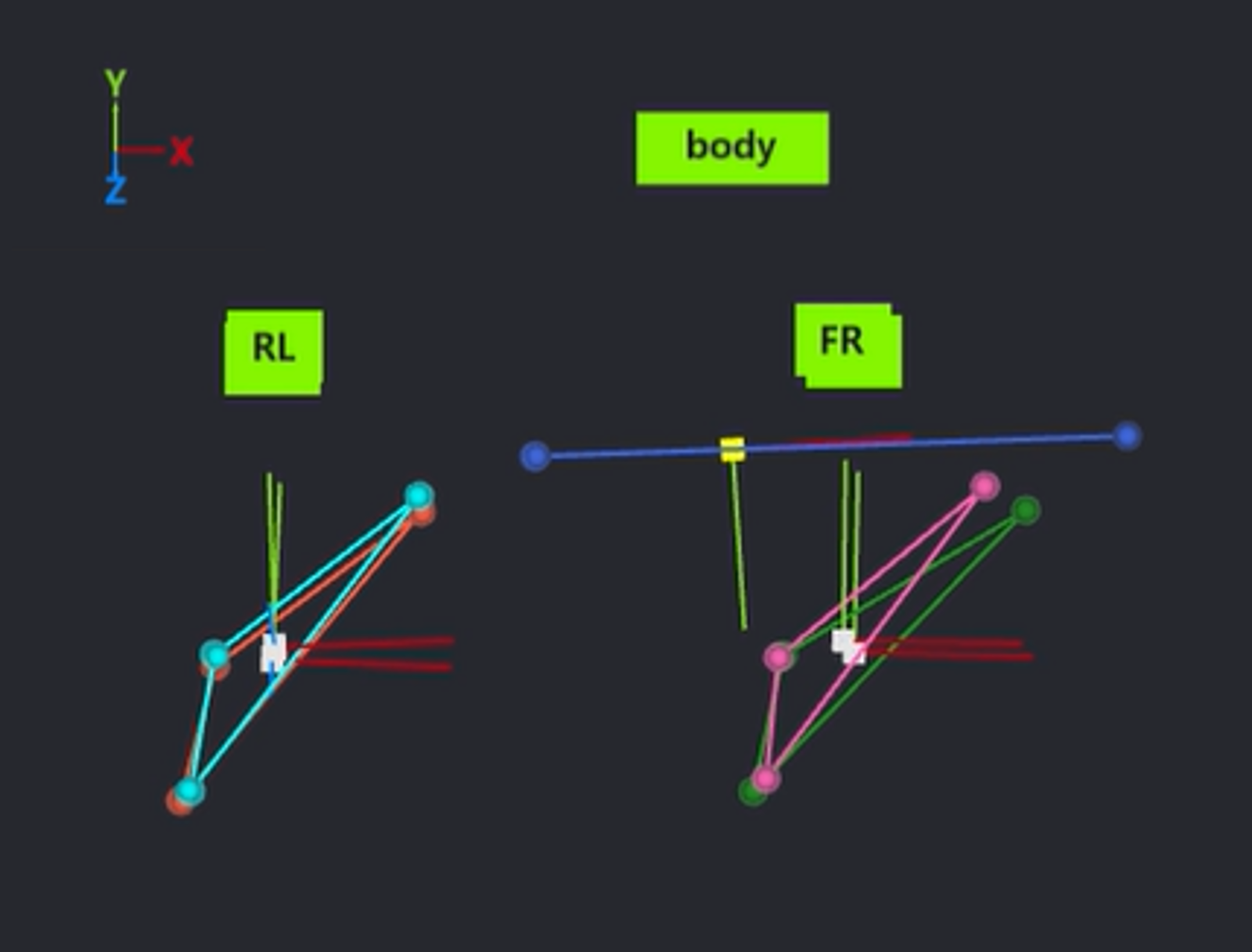}
        \subcaption*{(B)} 
        \label{fig:mocap_interface}
    \end{subfigure}
    
    \caption{The physical experimental setup. (A) The attached tracking balls (each rigid body circled in one color). (B) The established Unitree Go2 model in the Luster FZMotion motion capture system.}
    \label{fig:real_setup}
\end{figure}

\begin{figure*}[ht!] 
  \centering
  \includegraphics[width=0.98\textwidth]{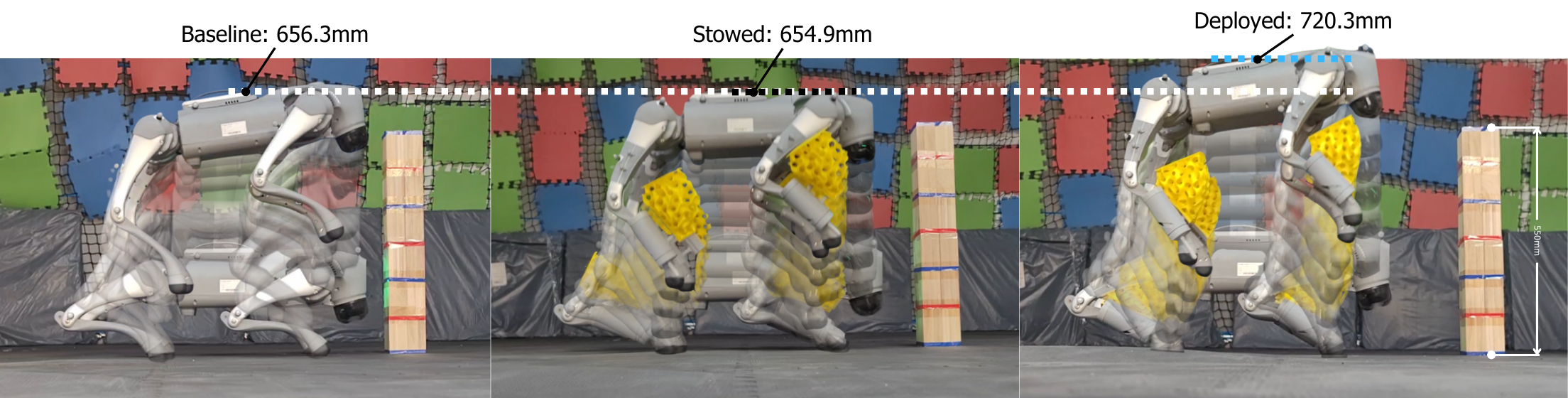} 
  
  \caption{Composite time-lapse photography of the vertical jumping experiments. 
  \textbf{(Left)} The \textit{Baseline} group (Unitree Go2 without payload) reaching a peak height of 656.3 mm. 
  \textbf{(Middle)} The \textit{Stowed} group (passive mass) reaching 654.9 mm, demonstrating negligible parasitic effect (0.33\% reduction) from the added mass. 
  \textbf{(Right)} The \textit{Deployed} group utilizing the PEBA lattice energy release, achieving a peak height of 720.3 mm, a 17.1\% increase over the baseline. 
  The white dashed reference line indicates the baseline peak height.}
  
  \label{fig:jump_comparison} 
\end{figure*}

\subsection{Experimental Setup}
Experiments were conducted on a physical Unitree Go2 robot. The compliant structure components were 3D printed using Arkema Pebax\textsuperscript{\textregistered} 3533 SP 01. All motions were tracked using a high-precision motion capture system (Luster FZMotion), providing ground-truth performance validation with 0.1 mm skeleton point tracking. 
To capture the robot's kinematics, a total of 15 MoCap tracking balls were attached to the system. Three markers were placed on the trunk (one on the anterior dorsal side and two on the lateral sides), and three were distributed along each thigh (proximal, middle, and distal)(Fig. \ref{fig:real_setup}A). This three-marker configuration defines a unique rigid body for each segment, enabling the motion capture system to resolve the full 6-DoF pose of every single component(Fig. \ref{fig:real_setup}B).

\subsection{Vertical Jumping Performance Validation}

To evaluate the robot's jumping capability with the compliant mechanism, three sets of vertical jumping experiments were conducted (Fig. \ref{fig:jump_comparison}):

\begin{enumerate}
    \item \textbf{Baseline:} The Unitree Go2 robot jumping without the added structure (Fig. \ref{fig:jump_comparison}, Left).
    \item \textbf{Stowed Group:} The robot with the SSCM attached but in the retracted state. This control group isolates the influence of the mechanism's additional weight on jumping performance (Fig. \ref{fig:jump_comparison}, Middle).
    \item \textbf{Deployed Group:} The robot with the module deployed, utilizing the combined effect of motor torque and the structure's elastic energy release (Fig. \ref{fig:jump_comparison}, Right).
\end{enumerate}



\begin{figure}[t]
\centering
 \includegraphics[width=0.7\columnwidth]{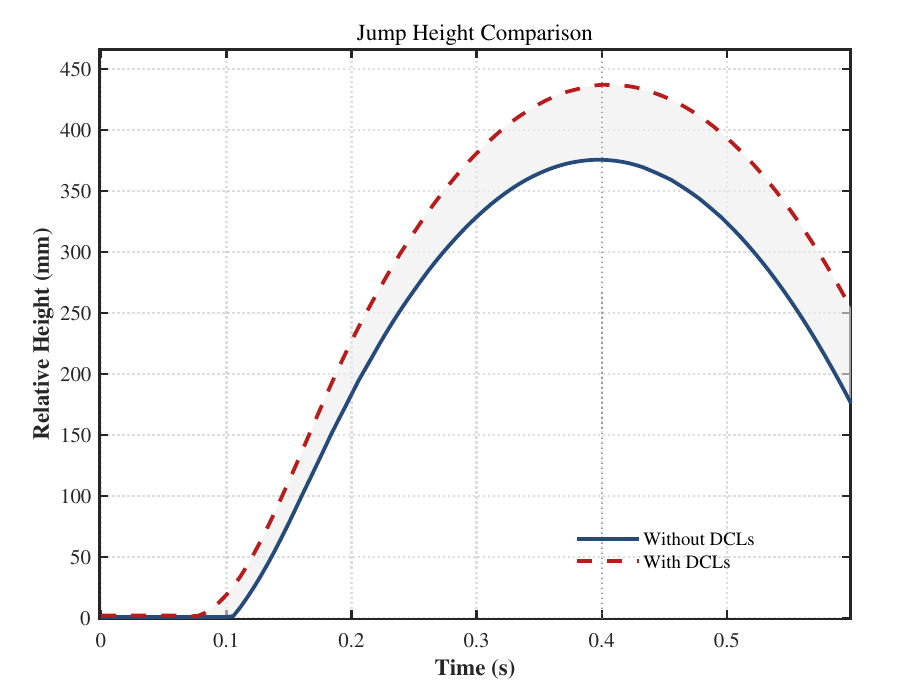}
\caption{Effective jump height comparison from motion capture data.}
\label{fig:JumpHeightComparison}
\end{figure}

To ensure a consistent comparison under mechanical constraints, the squat height was standardized across all trials at $H_{\text{base}} = 283.1\,\text{mm}$. Furthermore, to decouple the confounding influence of battery voltage sag on the motor's torque-speed envelope, the electrical energy state was strictly standardized. The robot's battery was maintained at a full charge state ($>95\%$ SoC) prior to each trial group to ensure consistent bus voltage availability for peak power output. The effective jump height is defined as $\Delta H = H_{\max} - H_{\text{base}}$. To mitigate experimental variance and ensure statistical reliability, five trials ($N=5$) were conducted for each experimental scenario (Baseline, Stowed, and Deployed). The results presented in this study represent the mean values derived from these trials. To quantify the enhancement, we define the \textbf{Relative Performance Change ($\delta$)} compared to the Baseline group:

\begin{equation}
    \delta = \frac{\Delta H - \Delta H_{\text{baseline}}}{\Delta H_{\text{baseline}}} \times 100\%
    \label{eq:jump_calc}
\end{equation}


The experimental results are summarized in Table \ref{tab:jump_results} and Fig. \ref{fig:JumpHeightComparison}. The \textbf{Baseline} group achieved an effective jump height of $\Delta H = 373.1\,\text{mm}$. The \textbf{Stowed} group achieved $\Delta H = 371.7\,\text{mm}$. Comparing these two, the added weight caused a negligible relative performance reduction of only $0.4\%$ ($\delta = -0.4\%$), indicating that the mass penalty of the structure is minimal.

In contrast, the \textbf{Deployed} group reached an effective jump height of $\Delta H = 437.1\,\text{mm}$. Under the combined action of the mechanism, the jump height increased by $64.0\,\text{mm}$ compared to the baseline. This corresponds to a significant \textbf{relative performance improvement of $17.1\%$} ($\delta = +17.1\%$).




\begin{table}[h]
\centering

\footnotesize 

\begin{tabular}{l c c c}
\hline
\textbf{\makecell[l]{Experimental\\Group}} & 
\textbf{\makecell{Max Height\\($H_{\max}$)}} & 
\textbf{\makecell{Eff. Jump Height\\($\Delta H$)}} & 
\textbf{\makecell{Relative Change\\(vs. Baseline)}} \\
\hline
\rule{0pt}{2.5ex}
Baseline & 656.3 mm & 373.1 mm & N/A \\
Stowed   & 654.9 mm & 371.7 mm & $-$0.4\% \\
\textbf{Deployed} & \textbf{720.3 mm} & \textbf{437.1 mm} & \textbf{+17.1\%} \\
\hline
\end{tabular}
\caption{Vertical Jumping Experimental Results (Mean values over $N=5$ trials)}
\label{tab:jump_results}
\end{table}

\section{CONCLUSIONS}

In this paper, we designed, modeled, simulated, and experimentally validated a novel, Deployable Compliant Leg (DCL) for quadruped robots. By combining a 3D-printed PEBA-based, Gyroid-filled structure with an analytical energy model and a segmented PD control strategy, our system demonstrated a significant and quantifiable \SI{17.1}{\percent} increase in vertical leaping height.

We have successfully shown that this system enhances explosive performance while, thanks to its deployable mechanism, not compromising nominal locomotion. 
Our FEA-driven design framework, which links analytical modeling (Eq. \ref{eq:poly_fit}) and FEA parameterization, provides a robust methodology for future designs.
Currently, the deployment is manually triggered. Future work will focus on automating this process, potentially through a passive mechanical latch triggered by a specific joint configuration (e.g., deep squatting) or a micro-actuator controlled by the robot's higher-level state machine.
In the future, we will deploy advanced control strategies, such as Model Predictive Control (MPC)\cite{11245991} or Reinforcement Learning (RL) \cite{cheng2022improving}, to fully exploit the complex, non-linear dynamics of this compliant system and further improve jump performance.

\section*{ACKNOWLEDGMENT}

This work was supported by the Summer Undergraduate Research Fellowships (SURF) programme at Xi'an Jiaotong-Liverpool University (XJTLU), Project No. SURF-2025-0179, and the Innovation Factory (IF) at XJTLU Entrepreneur College (Taicang), XJTLU, Project No. IF-36.







% ...


\bibliographystyle{IEEEtran}
\bibliography{references}
\end{document}